\documentclass{article}

\usepackage{fullpage}
\usepackage{algorithm}
\usepackage{algorithmic}
\usepackage{amsmath}
\usepackage{booktabs}
\usepackage{amsfonts}
\usepackage{amssymb}
\usepackage{preamble}
\usepackage{amsthm}
\usepackage{comment}
\usepackage{graphicx}
\usepackage[numbers,sort&compress]{natbib}
\usepackage{hyperref}

\definecolor{mydarkblue}{rgb}{0,0.08,0.45}
\hypersetup{ %
pdftitle={},
pdfkeywords={},
pdfborder=0 0 0,
pdfpagemode=UseNone,
colorlinks=true,
linkcolor=mydarkblue,
citecolor=mydarkblue,
filecolor=mydarkblue,
urlcolor=mydarkblue,
}

\usepackage{caption}

\theoremstyle{definition}
\newtheorem{problem}{Problem}
\newtheorem{remark}{Remark}
\newtheorem{definition}{Definition}
\newcommand{\eqrefrange}[2]{(\ref{#1})-(\ref{#2})}
\newcommand{\refrange}[2]{\ref{#1}-\ref{#2}}

\begin{document}

\title{Unifying Post-hoc Explanations of Knowledge Graph Completions}

\author{
  Alessandro Lonardi \and
  Samy Badreddine \and
  Tarek R. Besold \and
  Pablo Sánchez-Martín\
}
\date{
  Sony AI, Barcelona, Spain \\[1em]
  \small
  \texttt{alessandro.lonardi@sony.com} \quad
  \texttt{samy.badreddine@sony.com} \\
  \texttt{tarek.besold@sony.com} \quad
  \texttt{pablo.sanchez2@sony.com}
}


\maketitle

\abstract{
    Knowledge Graphs organize information as entity-relation-entity triples, enabling machine learning models to predict plausible missing triples in a task known as Knowledge Graph Completion (KGC).
Post-hoc explainability for KGC addresses the problem of identifying which triples most influence the predictions of machine learning models.
Currently, the field lacks formalization and consistent evaluations, hindering reproducibility and cross-study comparisons.
This paper argues for a unified taxonomy for post-hoc explainability in KGC.
First, we propose a characterization of post-hoc explanations via multi-objective optimization that unifies existing post-hoc explainability algorithms in KGC and the explanations they produce, balancing explanation effectiveness and conciseness.
Next, we examine improved evaluation protocols based on popular metrics, such as Mean Reciprocal Rank and Hits@$k$, through illustrative experiments.
Finally, we stress the importance of interpretability as the ability of explanations to address queries meaningful to end users.
By unifying methods and discussing evaluation standards, this work puts forward a case for more reproducible and impactful research in KGC explainability.
}

\section{Introduction}
\label{sec: introduction}

A Knowledge Graph (KG) is a collection of triples where directed relations $r$ endowed with predicates link two nodes, a subject $s$, and an object $o$. Nodes are commonly called entities since they possess semantics. They can be real-world objects or abstract concepts, and relations denote their connections. For example, a triple may be $(s,r,o) = ($\text{France}$,$ \text{located$\_$in}$,$ \text{Europe}$)$, or $(s,r,o) = ($\text{Insulin}$,$ \text{treats}$,$ \text{Diabetes}$)$.

Since their early days, KGs have stored massive amounts of data: the Freebase KG has hundreds of millions of triples \cite{bollacker2008freebase}. This made KGs a rich resource for applications from Natural Language Processing \cite{zhou2020kbert} to drug discovery \cite{lin2020kgnn}. Since many KGs are incomplete \cite{nickel2015review}, extensive research has been conducted in Knowledge Graph Completion ($\kgc$), an inference problem consisting of predicting missing information in a $\kgtext$ \cite{rossi2021knowledgereview}. 
If $(s,r,\text{?})$ $ = $ $($\text{France}$, $ \text{located$\_$in}$,$ ?$)$ or $(\text{?},r,o)$ $ = $ $($\text{?}$,$ \text{treats}$,$ \text{Diabetes}$)$ were missing, KGC would consist of {object completion}, inferring $\hat{o} =$ Europe, or {subject completion}, inferring $\hat{s} = $ Insulin.
Typically, KGC is carried out by Knowledge Graph Embedding $(\kge)$ models \cite{bordes2013translating,dettmers2018convolutional,yang2014embedding,trouillon2016complex,sun2018rotate}, which learn low-dimensional embeddings of entities and relations to give a high score to plausible inferred completions.
What is notable is that $\kgc$, and in particular post-hoc explainability for KGC, is increasingly used to support decision-making. For example, post-hoc explanations and predictions from biomedical KGs can assist scientists in their laboratory experiments \cite{sudhahar2024experimentally,huang2024foundation}. Despite the potential of post-hoc explanations in $\kgc$, comparatively little attention has been paid to formalizing post-hoc explanations to unify the outputs of different algorithms and to discuss metrics that assess explanation quality.

This work focuses on post-hoc explainability for KGC, which typically addresses the questions: \textbf{Q1.} {Which training triples most affect the score of a KGE model's prediction?}
\textbf{Q2.} {How much do they change the score of a prediction?} Such training triples are referred to as its post-hoc explanations\footnote{We use the terms \enquote{post-hoc explanation} and \enquote{explanation} interchangeably, while referring to post-hoc explanations.}. Finding them helps to draw a direct connection between data and predictions, a task otherwise impossible with a KGE model alone.

Commonly, two explanation types are discussed: \emph{necessary} and \emph{sufficient} \cite{watson2021local}. 
Necessary explanations act as subtractive counterfactuals for a prediction, i.e., their removal from the training set lowers the prediction score \cite{byrne2019counterfactuals}. 
Sufficient explanations are triples that alone suffice to enable a prediction independent of the rest of the training set. They are found with Graph Neural Networks (GNNs) used to distill the KGE model's scores \cite{ma2024kgexplainer,chang2024path,zhang2023page}, as a retraining of a KGE model requires more context than the explanation alone.
Alternatively, they are found by proxy as additive counterfactuals, i.e., by linking them to new entities and evaluating the increase in the prediction score for those entities \cite{rossi2022explaining,byrne2019counterfactuals}.

\paragraph{Research gaps.}
Although critical to aligning black-box models with user needs, post-hoc explanations in KGC lack a structured framework for formalizing and organizing them.

Beyond the necessary-sufficient distinction, structural constraints are imposed on explanations, like limiting them to a single triple \cite{pezeshkpour2019investigating,betz2022adversarial,bhardwaj2021adversarial} or sets of triples \cite{rossi2022explaining} featuring an entity from the prediction, or within $n$ hops from the prediction \cite{zhang2019data}. 
Algorithmic constraints make models efficient but introduce considerations for fair comparison.

\begin{figure}[t]
    \centering
    \includegraphics[width=0.58\columnwidth]{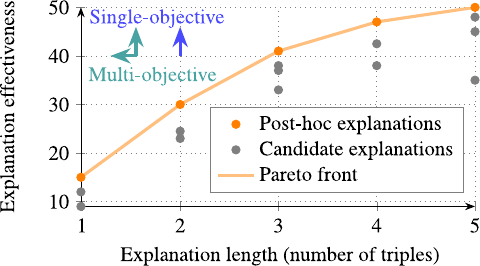}
    \caption{Multi-objective problem for post-hoc explanations. Multiple optimal explanations (orange) lie on the Pareto front. More restrictive algorithms solve a single-objective problem (blue) with a fixed explanation length, limiting the search space. More flexible algorithms solve a multi-objective problem (teal), varying both length and effectiveness. Axes follow the example in Sec. \ref{sec: nec expl}.}
    \label{fig: pareto-intro}
\end{figure}

First, they blur the definition of a post-hoc explanation, which should be consistent across studies. A post-hoc explanation pertains to KG data, while constraints are design choices of explainability methods.
Second, they can lead to incomplete comparisons between findings. 
Explainability algorithms often produce explanations of varying lengths (number of triples).
However, studies tend to compare results by focusing only on explanation effectiveness, overlooking the interplay between explanation length and effectiveness within a multi-objective evaluation framework.
See Fig. \ref{fig: pareto-intro} for a problem sketch.

These observations hold for necessary and sufficient explanations alike, with the latter presenting additional challenges, as learning KGEs from them is a difficult task that depends on several other factors.

In addition, many explainability algorithms \cite{pezeshkpour2019investigating,bhardwaj2021adversarial,betz2022adversarial} study the addition of unobserved training triples to alter prediction. They frame the problem as an adversarial task, one that can be enriched by a complementary perspective from generative models that impose a principled mechanism for sampling triples and supporting plausible explanations.

Finally, experiments can benefit from more robust evaluations. 
Current approaches often focus on performance metrics such as MRR (the Mean Reciprocal Rank on a set of test triples) and Hits@$k$ (the number of test triples ranking within the top $k$).
While these aggregated metrics provide valuable benchmarks, they can be hard to interpret in isolation and may not fully capture a method's effectiveness.
Moreover, evaluations can benefit from studying explanations through the lens of interpretability, intended as their capacity to answer user queries about model predictions.

\paragraph{Contributions.}
We argue that the lack of a rigorous and unified taxonomy for post-hoc explanations in KGC undermines the field's rigor, hinders cross-study comparability, and impairs reproducibility. 
To improve current standards, explanations should be both effective and concise, evaluated using complementary metrics, and interpretable for end users.
We organize our contributions as follows.

\begin{itemize}
    \item[\textbf{C1.}] We define post-hoc explanations in KGC as solutions to a multi-objective problem where explanation length, measuring its complexity, and effectiveness are jointly optimized (Sec. \ref{sec: post-hoc explanations}). 
    This definition unifies necessary (Sec. \ref{sec: nec expl}) and sufficient (Sec. \ref{sec: suff expl}) explanations, defined by tailoring the problem to specific effectiveness objectives.
    We also connect additive triples unobserved in the KG to explainability by formalizing \emph{latent} explanations (Sec. \ref{sec: lat expl}), which are unobserved triples sampled from a generative model.
    Our definitions do not depend on algorithmic constraints, enabling rigorous discussions.
    Within this classification, we argue that explanations should be evaluated with Pareto efficiency (Sec. \ref{sec: post-hoc explanations}). 
    Specifically, an explanation is not better than another if its higher effectiveness comes at the cost of greater length.

    \item[\textbf{C2.}] We discuss strategies to improve explainability experiments in KGC (Sec. \ref{sec: experiments}). To this end, we test four algorithms: Kelpie \cite{rossi2022explaining}, {\datapoisoning} \cite{zhang2019data}, CRIAGE \cite{pezeshkpour2019investigating}, and the {\anyburl} \cite{betz2022adversarial}, on ComplEx \cite{trouillon2016complex} and FB15k-237 \cite{toutanova2015observed}, a prominent KGE model and dataset for KGC. We publish our code at \texttt{\href{https://github.com/SonyResearch/unified-posthoc-explanations-kgc}{https://github.com/SonyResearch/unified-posthoc-explanations-kgc}}.

    \item[\textbf{C3.}] We advocate for a \enquote{sober look} \cite{bordt2024statistic} on explanations in KGC (Sec. \ref{sec: interpretability}). Explainability algorithms for KGC are agnostic to the meaning of explanations, i.e., they measure their effectiveness to sway predictions without knowing their semantics. This is a critical point to address, as explanations should assist humans in answering interpretative questions and tie \textbf{Q1-Q2} to user-relevant, real-world applications.
\end{itemize}

\section{Background}
\label{sec: background}

\subsection{Related Works}

Post-hoc explainability for black-box models has been extensively studied in recent years \cite{guidotti2024counterfactual,arrieta2020explainabile,adadi2018peeking,bodria2023benchmarking,nauta2023from,bugueno2024graph,rajabi2024knowledge}. 
Existing surveys encompass several machine learning tasks and applications, offering broad applicability but lacking domain-specific depth to tailor discussions to KGC.

In KGC, researchers mainly focused on developing explainability algorithms.
A popular approach to find explanations is to estimate the change in a KGE model's score upon altering the training set, e.g., with its Taylor approximation \cite{pezeshkpour2019investigating,zhang2019data}, with 
Instance Attribution measures \cite{gu2023iae,bhardwaj2021adversarial}, or relearning KGEs
\cite{rossi2022explaining,barile2025addititve}.
An alternative is to mine logical rules from the KG to retrieve paths that enable predictions \cite{betz2022adversarial,arakelyan2021complex,sadeghian2019drum,zhang2010iteratively}.
Recently, GNNs distilling KGE models' scores successfully extracted explanations as subgraphs with loose constraints on their structure
\cite{ma2024kgexplainer,chang2024path,huang2024foundation,zhang2023page}.

Despite many methodological advancements, definitions of post-hoc explanations in KGC are underdeveloped and fragmented. 
To the best of our knowledge, this is the first work to formalize a taxonomy by combining existing work, while also advocating for shared and more rigorous evaluation protocols for post-hoc explanations in KGC.

In this work, we focus on local explanations, i.e., training triples that account for individual predictions. We note, however, that global methods provide a complementary and valuable perspective on explainability, e.g., by finding explanations for clusters of entities \cite{gad2020ex}, or learning interpretable models that behave like the originally trained one \cite{sanchez2015towards,gusmao2018interpreting}.

\subsection{KGs and KGC}

We denote a KG as $\kg = \{ (s,r,o)\}$, where each triple consists of two entities, the subject $s$ and the object $o$, linked by a directed relation $r$.
Let $\Omega := \entities \times \relations \times \entities$ be the set of all possible triples that can be built using the entity set $\entities$ and the relation set $\mathcal{R}$.
A KG is a directed multigraph (a directed graph with potentially multiple edges across nodes) $\kg \subseteq \Omega$.

{KGC consists of completing $\kg$ with plausible missing triples.}
KGE models are SOTA for KGC. Assume that $\kg$ is partitioned with the disjoint sets $\kg = \kg_\train \cup \kg_\valid \cup \kg_\test$. 
A $\kge$ model with parameters $\boldsymbol{\theta}$ is trained on $\kg_\train$ to learn low-dimensional embeddings of entities and relations and a score function $\scorefn(\subject, \relation, \object) \in \mathbb{R}$.
The score function assigns higher values to plausible triples, and the learned embeddings enable inference, i.e., completion. From here on, we discuss object completion with analogous arguments valid for subject completion.
The quality of a KGE model is commonly assessed by ranking the score of test triples $(\subject,\relation, \object) \in \kg_{\text{test}}$.
In formulas, this can be done with the rank
\begin{equation}
    \label{eq: ranking def}
    \rank(\subject, \relation, \object) := 1 + \sum_{\entity \in \entities'} \mathbf{I} [ \scorefn(\subject, \relation, \entity) > \scorefn(\subject, \relation, \object) ] \,.
\end{equation}
Here, \( \mathbf{I}[\cdot] \) is the indicator function, equal to 1 if the condition in square brackets is true and 0 otherwise. The set $\entities' := \{ \entity \in \entities \, | \, (\subject, \relation, \entity) \notin \kg \}$ filters the rank by only including object entities that do not form KG triples, ensuring accurate evaluations across multiple KGE models \cite{rossi2021knowledgereview}.

In the following, we discuss post-hoc explanations for predictions $(s_x, r_x, \hat{o}_x)$, i.e., triples completed by the KGE model.
We add the subscript $x$ to distinguish them from the rest of the KG clearly.

\section{Unifying Post-hoc Explanations}
\label{sec: post-hoc explanations}

We define post-hoc explanations as solutions of a multi-objective optimization problem that, informally, finds the most \emph{concise} and most \emph{effective} explanation.

First, we introduce the search space $\mathcal{X} := \{ (s,r,o) \in \Omega \, | \, C_k(s,r,o) = \text{True} \; \forall k = 1,\dots, K \}$ where candidate explanations, i.e., sets of triples $X$, lie. This is a subset of all possible triples in $\Omega$ subject to $K$ Boolean constraints $C_k$ involving entities and relations. For instance, to restrict $\mathcal{X} = \kg$, we write $C_1(s,r,o) = (s,r,o) \in \kg$, $K = 1$.
Next, we introduce $\mathsf{F}$, a retraining algorithm for the KGE model. 
Given a candidate explanation $X$ and the training set $\kg_\train$, $\mathsf{F}$ learns $\boldsymbol{\theta'}$, thus it updates KGE score function to $\scorefnprime$.
As we discuss later, the implementation of $\mathsf{F}$ changes with the type of explanation. 
For example, for necessary explanations, $\mathsf{F}$ removes $X$ from $\kg_\train$ and retrains the KGE model on $\kg_\train \setminus X$. 
Lastly, we define a function $\Psi(\expl ; \subject_x, \relation_x, \hat{\object}_x, \mathsf{F}, \boldsymbol{\theta}) \in \mathbb{R}$ measuring the effectiveness of $X$ on a KGE prediction $(s_x,r_x,\hat{o}_x)$, once the KGE model $\boldsymbol{\theta}$ is retrained with $\mathsf{F}$.
This allows us to define post-hoc explanations as jointly maximizing the \emph{effectiveness} of the explanation through $\Psi$---a problem-dependent measure of how it impacts the prediction (e.g., how much it moves the prediction's rank)---while keeping it as \emph{concise} as possible through $|X|$, the number of triples it uses.

\begin{problem}
\label{prob: multiobjective}
Given a prediction $(\subject_x, \relation_x, \hat{\object}_x)$ inferred by a $\kge$ model, post-hoc explanations $\explstar$ are solutions to the multi-objective optimization problem
\begin{align}
    \label{eq: explanation problem 1}
    \text{Minimize} \quad & \lvert \expl \rvert \\
    \label{eq: explanation problem 2}
    \text{Maximize} \quad & \Psi(\expl ; \subject_x, \relation_x, \hat{\object}_x, \mathsf{F}, \boldsymbol{\theta}) \\
    \label{eq: explanation problem 3}
    \text{s.t.} \quad & \expl \subseteq \mathcal{X} \setminus \emptyset \,.
\end{align}
\end{problem}

\begin{remark}
    Typically, explainability algorithms do not fully retrain the KGE model with $\mathsf{F}$ because this is computationally expensive. 
    Instead, they approximate the impact of removing an explanation using disparate retraining proxies.
    Our classification naturally accommodates these approximations. 
    To unify them, we focus on Prob. \ref{prob: multiobjective}, which serves as the fundamental formulation of the explainability task.
\end{remark}

We now discuss key aspects of the formulation in Prob. \ref{prob: multiobjective}.
First, the problem is independent of any algorithmic implementation, and therefore, it is suitable for modeling post-hoc explanations in a standardized manner.
Second, we argue that explanations ought to solve a multi-objective problem in which potentially contrasting, interdependent objectives, e.g., explanation length and effectiveness, are simultaneously optimized.

More generally, explanations should be as concise and effective as possible. We measure conciseness with the cardinality of $X$ as already done in the literature \cite{ma2024kgexplainer,rossi2022explaining,chang2024path}, though one can adopt other measures in a similar vein.
Concise explanations are desirable as they are easier to understand \cite{nauta2023from,bhatt2020evaluating,cui2019an,miller2019explanation}. 
High effectiveness is important to retain prediction information and faithfulness \cite{Agarwal2022OpenXAITA}. While we use rank change for a predicted triple as an instrumental measure of effectiveness, other context-dependent choices are valid.
For example, one could consider adding a ``soft preference'' for path-based explanations in $\Psi$. An interesting direction is also to consider multiple effectiveness functions $\Psi_i$ encoding contrastive objectives, e.g., rank change and preference for path-based explanations (leading to an $n$-dimensional Pareto front). We leave this to future work.

Algorithms in the literature often rely on single-objective optimization, with $|X|$ fixed a priori \cite{betz2022adversarial,pezeshkpour2019investigating,zhang2019data,bhardwaj2021adversarial}, thus limiting the diversity of solutions.
Recent approaches \cite{rossi2022explaining,ma2024kgexplainer,chang2024path} explore flexible strategies that balance length and effectiveness, offering promising directions for finding diverse solutions.
Because of the multi-objective nature of Prob. \ref{prob: multiobjective}, the natural approach to evaluate explanations is through Pareto efficiency. That is, an explanation $X_1$ outperforms $X_2$ if $X_1 \leq X_2$ and $\Psi(X_1) \geq \Psi(X_2)$, with at least one strict inequality. 
As shown in Fig. \ref{fig: pareto-intro}, a long explanation, even if more effective, may lie on the Pareto front. Thus, it does not exhibit clear dominance over a less effective but shorter one.

\subsection{Necessary Explanations}
\label{sec: nec expl}

We define necessary explanations by formalizing a notion that is commonly operationalized in algorithms.

\begin{definition}[\emph{Necessary Explanation}]
\label{def: nec expl}
A necessary post-hoc explanation $\explstarn$ for a prediction $(\subject_x, \relation_x, \hat{\object}_x)$ with rank $\rank(\subject_x, \relation_x, \hat{\object}_x) < \lvert \entities' \rvert$ is the smallest non-empty set of training triples that, if removed from $\kg_\train$ and upon retraining the $\kge$ model on $\kg_\train \setminus \explstarn$, maximally increases $\rank(\subject_x, \relation_x, \hat{\object}_x)$.
\end{definition}

Increasing the rank means worsening the prediction score.
Crucially, $\explstarn$ in Def. \ref{def: nec expl} can be expressed as solutions of Prob. \ref{prob: multiobjective}. For that, \textbf{(i)} we restrict the search space to $\mathcal{X} = \kg_\train$ with $C_1(s,r,o) = (s,r,o) \in \kg_\train$. \textbf{(ii)} $\mathsf{F}$ removes $X$ from $\kg_\train$ and retrains the KGE model on $\kg_\train \setminus X$. \textbf{(iii)} We use $\Psi$ to measure rank change as in \citet{rossi2022explaining} (omitting the arguments of $\Psi$ for brevity)
\begin{equation}
\label{eq: score function nec}
\Psi = {\rankprime(\subject_x, \relation_x, \hat{\object}_x) - \rank(\subject_x, \relation_x, \hat{\object}_x)} \,.
\end{equation}

\begin{figure}[t]
    \centering
    \includegraphics[width=0.58\columnwidth]{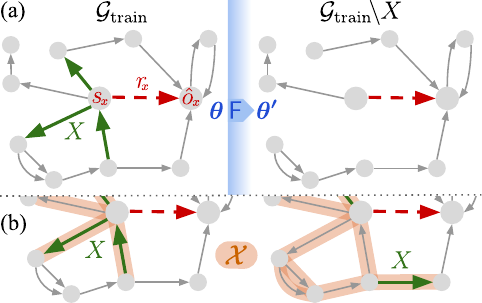}
    \caption{(a) A candidate necessary explanation $X$ (green solid) for $(s_x, r_x, \hat{o}_x)$ (red dashed) is removed by $\mathsf{F}$ from $\kg_\train$, which then retrains the KGE model on $\kg_\train \setminus X$. (b) A larger search space $\mathcal{X}$ (orange) may yield shorter explanations. Left: $\mathcal{X}$ is made of neighboring triple with $(s_x, r_x, \hat{o}_x)$ and $|X|=3$. Right: $\mathcal{X}$ is made of triples at 1-hop distance from $(s_x, r_x, \hat{o}_x)$ and $|X|=2$.}
    \label{fig: nec-expl-fig}
\end{figure}

In Fig. \ref{fig: nec-expl-fig}(a), we draw a candidate necessary explanation $X$ removed from the training set for KGE model retraining.

\begin{remark}
\label{remark: message passing}
For many popular KGE models, e.g., ComplEx \cite{trouillon2016complex}, ConvE \cite{dettmers2018convolutional}, TransE \cite{bordes2013translating}, DistMult \cite{yang2014embedding}, RotatE \cite{sun2018rotate}, it may be a reasonable heuristic to restrict the search space $\mathcal{X}$ to $\text{WCC}(s_x)$. $\text{WCC}(s_x)$ is the unique connected subgraph of $\kg_\train$ containing $s_x$. This restriction becomes exact when (i) model parameters are specific to each connected component, (ii) relation embeddings are fixed, or (iii) the chosen architecture $\mathsf{F}$ prevents parameter sharing across connected components. In such cases, triples in other connected components do not influence embeddings within $\text{WCC}(s_x)$.
\end{remark}

Instantiating Prob. \ref{prob: multiobjective} exemplifies how optimal explanations correspond to multiple sets of triples on the Pareto front of Eqs. \eqrefrange{eq: explanation problem 1}{eq: explanation problem 3}.
Take a top-ranked prediction with $\rank(s_x,r_x,\hat{o}_x) = 1$ and measure the effectiveness of its necessary explanations as in Eq. \eqref{eq: score function nec}. If we find two explanations of length $|{X^\star_{n,1}}| = 1$ and $|{X^\star_{n,2}}| = 2$, with effectiveness $\Psi({X^\star_{n,1}}) = 15$ ($\rankprime(s,r,\hat{o}) = 16$) and $\Psi({X^\star_{n,2}}) = 30$ ($\rankprime(s,r,\hat{o}) = 31$), we cannot deem one better than the other in the Pareto sense. Both lie on the Pareto front, i.e., neither surpasses the other on both optimization objectives. 

Furthermore, Def. \ref{def: nec expl} does not restrict Prob. \ref{prob: multiobjective} to a \emph{specific type} of necessary explanation, e.g., single triples. Instead, it decouples the definition of an explanation from the constraints of the algorithms extracting it.
The literature tends to overlook this distinction by focusing on algorithms rather than on formal discussions of explanations.
For example, several algorithms \cite{pezeshkpour2019investigating,betz2022adversarial,bhardwaj2021adversarial,zhang2019data} find explanations that are single triple sharing an entity with $(s_x,r_x,\hat{o}_x)$, solving a single-objective problem in the \enquote{effectiveness direction} at length 1 (see Fig. \ref{fig: pareto-intro}). \citet{zhang2019data} enlarge $\mathcal{X}$ to find $M$ triples at 1 hop from $(s_x,r_x,\hat{o}_x)$. As shown in Fig. \ref{fig: nec-expl-fig}(b), a bigger $\mathcal{X}$ may yield shorter yet effective explanations, with a better Pareto trade-off in Prob. \ref{prob: multiobjective}. More generally, \citet{rossi2021knowledgereview} build explanations of varying length. They restrict $\mathcal{X}$ with a top-$K$ policy filtering the neighbors of $(s_x, r_x, \hat{o}_x)$.

We summarize algorithmic constraints in Table \ref{tab: table algorithm}. Here, we collect algorithms that either solve a single-objective problem with a fixed explanation length or a multi-objective problem with a variable explanation length. Different restrictions are imposed on the search space of explanations, $\mathcal{X}$. When convenient, we express these constraints so that they can be interpreted as Boolean functions $C_k(s,r,o)$.  We use \enquote{N} and \enquote{S} to indicate the algorithms' capacity to extract necessary and sufficient explanations, respectively. The label \enquote{A} is used for algorithms that add unseen triples to the KG through adversarial attacks to alter predictions. We expand further on this point in Sec. \ref{sec: lat expl}. Finally, asterisks indicate implementations from \citet{rossi2022explaining}. In Appendix Sec. \ref{apx-sec: explainability algorithms}, we give details on the functioning of CRIAGE, Data Poisoning, Kelpie, and the AnyBURL explainer.

Finally, we note again that although Def. \ref{def: nec expl}, and in turn Eq. \ref{eq: score function nec}, instantiates and conciseness ($|X|$) and effectiveness ($\Psi$) with measures commonly regarded as standard reference choices, this instantiation primarily serves to ground the problem in a concrete formulation. In general, the objectives of Prob. \ref{prob: multiobjective} can be changed to accommodate context-dependent problems.

\begin{table*}[htpb]
\begin{center}
\resizebox{0.895\textwidth}{!}{%
\begin{tabular}{lcccc}
\toprule
Algorithm & Expl. length & $\mathcal{X}$ ($C_k(s,r,o)$)& N,  S, A & Description \\
\midrule
{\criage} & Fixed, $1$ & $o = \hat{o}_x$ & N, S*, A &  Sec. \ref{apx-ssec: criage} \\
{\instanceattribution} & Fixed, $1$ & $( s \vee o ) = \{ s_x, \hat{o}_x \}$ & N, A & --- \\
{\anyburl}& Fixed, $1$ & $( s \vee o ) = \{ s_x, \hat{o}_x \}$ & N, A & Sec. \ref{apx-ssec: anyburl explainer} \\
{Data Poisoning} & Fixed, $\geq 1$ & Case 1: $s = s_x$ / Case 2: 1-hop KG from $s_x$ & N, S*, A & Sec. \ref{apx-ssec: data poisonining} \\
{\kelpie} & Variable, $\leq 4$ & $s = s_x \vee o = s_x$ and Top-$K$ $= 10,20,30$ & N, S & Sec. \ref{apx-ssec: kelpie} \\
\bottomrule
\end{tabular}
}
\caption{Summary of explainability algorithms for KGC.}
\label{tab: table algorithm}
\end{center}
\end{table*}

\subsection{Sufficient Explanations}
\label{sec: suff expl}

A sufficient explanation is a set of training triples capturing the minimal yet sufficient information in the KG required to make a prediction.
In principle, a candidate explanation of a prediction $X$ would be sufficient if learning KGEs on $X$ alone yielded that same prediction. More precisely, we formalize its commonly operationalized definition.

\begin{definition}[\emph{Sufficient Explanation}]
\label{def: suf expl ideal}
A sufficient post-hoc explanation $\explstars$ for a prediction $(\subject_x, \relation_x, \hat{\object}_x)$ with rank $\rank(\subject_x, \relation_x, \hat{\object}_x)$ is the smallest non-empty set of training triples that, upon retraining the $\kge$ model on $\explstars$ alone, minimally perturbs $\rank(\subject_x, \relation_x, \hat{\object}_x)$.
\end{definition}

We also obtain $\explstars$ in Def. \ref{def: suf expl ideal} by instantiating Prob. \ref{prob: multiobjective}.
\textbf{(i)} We set $\mathcal{X} = \kg_\train$ with $C_1(s,r,o) = (s,r,o) \in \kg_\train$.
\textbf{(ii)} $\mathsf{F}$ retrains the KGE model only on a candidate explanation $X$, ignoring other triples in $\kg_\train$.
\textbf{(iii)} We measure the effectiveness of $X$ by how little it degrades the prediction score, i.e.,
\begin{equation}
    \Psi =  \rank(\subject_x, \relation_x, \hat{\object}_x) - \rankprime(\subject_x, \relation_x, \hat{\object}_x) \,.
\end{equation}

\begin{remark}
\label{remark: gnn}
Steps \textbf{(i)-(iii)} are conceptually analogous to those performed by \citet{ma2024kgexplainer}, with the important exception that they estimate the retraining $\mathsf{F}$ of the KGE model by distilling a GNN from it.
\end{remark}

There are some practical considerations related to Def. \ref{def: suf expl ideal}.
Step \textbf{ (ii)} is highly impractical for KGE models that learn embeddings from the whole KG. 
In fact, training a KGE model on a candidate sufficient explanation $X$ alone, especially if small, would hardly produce meaningful embeddings. 
GNNs are valuable architectures for overcoming this obstacle, but no panacea. For example, \citet{ma2024kgexplainer} anchor the KGEs as learned on the whole KG through knowledge distillation and selective freezing.
In particular, they freeze KGEs outside of \enquote{a vicinity of the prediction} (within 2 hops from $s_x$ and $\hat{o}_x$), and in step \textbf{(ii)}, they retrain on $X \cup X_{\text{ext}}$ instead of $X$. Here, $X_{\text{ext}}$ is a set of triples that connects $X$ to the frozen KG, ensuring that the retrained KGEs are correctly anchored in the context of the frozen ones.
This strategy suffers from data contamination, as the extracted $X_s^\star$ will not be independent of the KG as intended. On the upside, GNNs are flexible and generate explanations with loose structural constraints.

As an alternative proxy, \citet{rossi2022explaining} define sufficient explanations differently and do not require the retraining of step \textbf{(ii)} we described.
They connect candidate explanations $X$ to new entities and measure their effectiveness by \enquote{how well the prediction transfers} to said new entities. To formalize this, we introduce
\begin{definition}
\label{def: target triples}
A ``target set'' $\mathcal{C} \subseteq \entities \setminus \{ s_x \}$ is a set of entities such that the triples in $\kg_\mathcal{C} := \{(c,\relation_x, \hat{\object}_x) \, | \, c \in \mathcal{C} \}$ are not top-ranked, i.e., $\rank(c,\relation_x, \hat{\object}_x) > 1$ for all $(c, \relation_x, \hat{\object}_x) \in \kg_\mathcal{C}$.
\end{definition}

A sufficient explanation found by proxy aims to make triples in $\kg_\mathcal{C}$ more plausible. In particular,

\begin{definition}[\emph{$\mathcal{C}$-Sufficient Explanation}]
\label{def: suf expl}
    Given a set $\kg_\mathcal{C}$, a $\mathcal{C}$-sufficient post-hoc explanation $X^\star_\mathcal{C}$ for a prediction $(s_x,r_x,\hat{o}_x)$ is the smallest non-empty set of training triples containing $s_x$ that, if added to $\kg_\train$ after swapping $s_x$ with $c$, and upon retraining the $\kge$ model at each substitution, maximally decreases $\rank(c, \relation_x, \hat{\object}_x)$ averaged over $\mathcal{C}$.
\end{definition}

Again, we get $X^\star_{\mathcal{C}}$ in Def. \ref{def: suf expl} from Prob. \ref{prob: multiobjective}. \textbf{(i)} We set $\mathcal{X} = \kg_\train$ with $C_1(s,r,o) = (s,r,o) \in \kg_\train$ \textbf{(ii)} Now, $\mathsf{F}$ first adds $X$ to the training set as $\kg_\train \cup X(s_x \to c)$, where $s_x \to c$ indicates the $s_x$-to-$c$ entity substitution, then it retrains the KGE model. \textbf{(iii)} We may measure rank change similarly to \citet{rossi2022explaining}, i.e.,
\begin{equation}
\label{eq: score function suf}
\Psi = \frac{1}{\lvert \mathcal{C} \rvert} \sum_{c \in \mathcal{C}} {\rank(c, \relation_x, \hat{\object}_x) - \rankprime(c, \relation_x, \hat{\object}_x)} \,.
\end{equation}

Def. \ref{def: suf expl} carries a series of nuances, which are discussed here.

\begin{remark}
    \label{rem:wcc_suff}
    Because of the $s_x$-to-$c$ substitution, the search space for $\mathcal{C}$-sufficient explanations is automatically constrained to triples containing $s_x$, and it is not $\kg_\train$.
\end{remark}

\begin{remark}
    $\mathcal{C}$-sufficient explanations may increase the rank, i.e., return a negative summand in Eq. \eqref{eq: score function suf}, for some triples $(c, \relation_x, \hat{\object}_x)$. However, the average rank over $\mathcal{C}$ must decrease. Alternatives are also valid, e.g., requiring all ranks of triples in $\kg_\mathcal{C}$ to decrease.
\end{remark}

\begin{figure}[t]
    \centering
    \includegraphics[width=0.58\columnwidth]{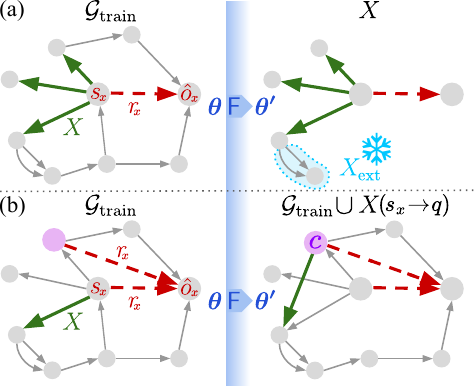}
    \caption{(a) A candidate sufficient explanation $X$ (green solid) for $(s_x, r_x, \hat{o}_x)$ (red dashed) is kept by $\mathsf{F}$, which then retrains the  KGE model on $X$ alone. As noted, such retraining is hard. Hence, $X_{\text{ext}}$ (blue shade) with frozen embeddings is used. (b)
    A candidate $\mathcal{C}$-sufficient explanation $X$ for $(s_x, r_x, \hat{o}_x)$ is found, and a ``target set'' $\mathcal{C} = \{c\}$ (pink) is defined.  $\mathsf{F}$ adds $X(s_x \to c)$ to $\kg_\train$ and retrains the KGE model to compute $\rankprime(c,r_x,\hat{o}_x)$.}
    \label{fig: suf-exp-fig}
\end{figure}

We illustrate the extraction and evaluation of a candidate sufficient and $\mathcal{C}$-sufficient explanation in Fig. \ref{fig: suf-exp-fig}.

Importantly, Def. \ref{def: suf expl} is a proxy departing from the standard notion of sufficiency in Def. \ref{def: suf expl ideal}. This approach renders the search for sufficient explanations for KGE models feasible through $\mathcal{C}$-sufficient explanations. 
%
%
However, it inevitably entails trade-offs that offer valuable opportunities for further exploration and are important for understanding its limitations.
First, the robustness of Def. \ref{def: suf expl} improves as $|\mathcal{C}|$ grows. 
In \citet{rossi2022explaining}, $\mathcal{C}$-sufficient evaluations are found by setting $|\mathcal{C}| = 10$ ($< 0.1\%$ of the entities in datasets like FB15k-237). 
While we recognize that extensive computations can quickly become challenging, they are crucial to improving experimental results.
Second, limiting the search space of $\mathcal{C}$-sufficient explanations as noted in Remark \ref{rem:wcc_suff} may yield results that are hard to interpret. 
For example, take a prediction $(s_x, r_x, \hat{o}_x)$  $=$ $($\text{Paris}$,$ \text{city\_in}$,$ \text{Europe}$)$ in a simple KG connecting $s_x$ to $\hat{o}_x$ with the path $($\text{Paris}$,$ \text{located\_in}$,$ \text{Île-de-France}$)$, $($\text{Île-de-France}$,$ \text{located\_in}$,$ \text{France}$)$, $($\text{France}$,$ \text{located\_in}$,$ \text{Europe}$)$ (see Appendix Sec. \ref{apx-sec: more suff expl} for a drawing). This path is a sufficient explanation in the standard sense, yet Def. \ref{def: suf expl} returns only its first triple with the $s_x$-to-$c$ swap.
%

\subsection{Latent Explanations}
\label{sec: lat expl}

We consider latent explanations to be additive triples unobserved by the KGE model during training, i.e., not in $\kg_\train$. 
While the addition of such triples has been studied in the context of adversarial attacks on predictions, we reframe this problem through a different lens.
In particular, we view $\kg$ as a sample drawn from a statistical ensemble on $\Omega$. In this context, latent explanations are plausible missing triples that a generative model can infer\footnote{We use a probabilistic model, as it yields a robust measure of closeness between samples via its likelihood. Non-probabilistic models can also work if a closeness measure is introduced.}. Incorporating them into the training set introduces new information to alter predictions.

We formalize this description as in \citet{loconte2023how}. Let $\mathbf{Y} \in \{0, 1\}^{|\entities| \times |\relations| \times |\entities|}$ be a tensor representing $\kg$ as $Y_{sro} = 1$ if $(s, r, o) \in \kg$ and $Y_{sro} = 0$ otherwise. Interpreting $Y_{sro}$ as a Bernoulli random variable yields the probability of sampling $\kg$ in $\Omega$,
that is
\begin{equation}
\label{eq: prob q}
q(\kg) = \prod_{(s,r,o) \in \Omega} p(Y_{sro} = \mathbf{I} [(s,r,o) \in \kg] \, | \, s, r, o) \,.
\end{equation}
We then define two types of latent explanations: \emph{positive} and \emph{negative}. We discuss the former, aiming at decreasing the rank of a prediction (increasing its score), in Def. \ref{def: pos lat expl}. The latter is presented in Appendix Sec. \ref{apx-sec: latent explanations}.
\begin{definition}[\emph{Positive Latent Explanation}]
\label{def: pos lat expl}
A positive latent post-hoc explanation $\explstarlplus$ for a prediction $(\subject_x, \relation_x, \hat{\object}_x)$ with rank $\rank(\subject_x, \relation_x, \hat{\object}_x) > 1$ is the smallest non-empty set of unobserved triples in $\Omega \setminus \kg_\train$ \enquote{likely with respect to $q$} that, if added to $\kg_\train$ and upon retraining the $\kge$ model on $\kg_\train \cup \explstarlplus$, maximally decreases $\rank(\subject_x, \relation_x, \hat{\object}_x)$.
\end{definition}

For example, consider an incomplete KG that expresses country-country and country-continent relationships. If $($\text{France}$,$ \text{neighbor\_of}$, $\text{ Germany}$)$, $($\text{France}$, $\text{ neighbor\_of}$, $\text{ Italy}$)$ were missing, sampling and adding them to the KG  could increase the score of the prediction $($\text{France}$,$ \text{ located\_in}$,$ \text {Europe}$)$.

\begin{remark}
    We do not specify the sampling scheme or formally define what \enquote{likely triples with respect to $q$} are, as these are design choices.  Still, Def. \ref{def: pos lat expl} captures the salient point of our construction: latent explanations are probabilistically plausible missing information of $\kg$.
\end{remark}

$\explstarlplus$ in Def. \ref{def: pos lat expl} is also a solution of Prob. \ref{prob: multiobjective}.
This requires to \textbf{(i)} restrict the search space to $\mathcal{X} = \Omega \setminus \kg_\train$ with $C_1(s,r,o) = (s,r,o) \notin \kg_\train$ and with constraints $C_k(s,r,o)$ for likely triples, e.g., $C_2(s,r,o) = \{(s,r,o) \in \Omega  \, | \, p(Y_{sro} = 1\, | \, s,r,o) \geq 1 - \varepsilon, \; \varepsilon \text{ \enquote{small}} \}$. \textbf{(ii)} $\mathsf{F}$ now adds $X$ to $\kg_\train$ and retrains the KGE model on $\kg_\train \cup X$. \textbf{(iii)} $\Psi$ can evaluate rank change if set as in Eq. \eqref{eq: score function nec}.

\begin{figure}[t]
    \centering
    \includegraphics[width=0.58\columnwidth]{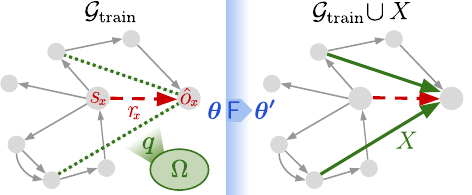}
    \caption{A candidate latent explanation $X$ (solid green) for $(s_x, r_x, \hat{o}_x)$ (red dashed) is drawn from $\Omega$ to fill two unobserved triples (dotted green).  $\mathsf{F}$ adds to $X$ to $\kg_\train$ and retrains the KGE model.}
    \label{fig: lat-exp-fig}
\end{figure}

We draw an illustration of the extraction and evaluation of a candidate latent explanation $X$ in Fig. \ref{fig: lat-exp-fig}.

The perspective of linking unobserved explanations to a generative model largely departs from most current research.
Only very recently, a similar approach to ours has been adopted in \citet{barile2025addititve}, where unseen triples are found with quotient graphs \cite{cebiric2019summarizing}. There, the authors also suggest that a solution to this task is to employ probabilistic generative algorithms such as GeKCs \cite{loconte2023how}, which makes sampling from Eq. \eqref{eq: prob q} tractable and, therefore, would enable the discovery of latent explanations.
In contrast, the literature \cite{pezeshkpour2019investigating,bhardwaj2021adversarial,betz2022adversarial,zhang2019data} commonly frames the addition of unobserved triples as an adversarial problem, but without a generative model (see Appendix Sec. \ref{apx-sec: latent explanations} for details). 
Remarkably, using a generative model, i.e., imposing a mechanism for sampling probable triples, holds the potential to uncover highly non-trivial plausible missing relations in a KG.
This could lead to substantial scientific impact, e.g., in biomedical studies \cite{huang2024foundation,sudhahar2024experimentally}.
Crucially, though, the insightfulness and truthfulness of latent explanations should always be assessed by a user, as is done for adversarial methods in, e.g., \citet{betz2022adversarial}.

\section{Experimental Evaluations}
\label{sec: experiments}

We are equipped with a taxonomy for explanations in KGC, so now we focus on experimental evaluations.
To accompany our discussion, we train ComplEx on FB15k-237. 
We find necessary explanations for 50 random top-ranked test triples in a set $\overline{\kg}$ with CRIAGE, Data Poisoning, the {\anyburl}, and Kelpie (extracting explanations with variable length and length 1; we write K1 in the latter case).
Details on experiments in Appendix Sec. \ref{apx-sec: experiments details}.

Typically, explainability algorithms are evaluated with
\begin{align}
    \label{eq: hits at k}
    \text{Hits@}k &:= |\{ (s,r,o) \in \overline{\mathcal{G}} \, | \, \rank(s,r,o) \leq k \}| \\
    \label{eq: mrr}
    \text{MRR} &:= \frac{1}{|\overline{\mathcal{G}}|}
     \sum_{(s,r,o) \in \overline{\mathcal{G}}} \frac{1}{\rank(s,r,o)}  \,.
\end{align}
Eq. \eqref{eq: hits at k} counts triples with ranks lower than $k$. Eq. \eqref{eq: mrr} is a softer alternative, with no cut at $k$ and weighing more low-ranked triples. These metrics effectively assess the performance of KGE models. 
However, we argue that if used in isolation or without proper consideration, they might make explainability results hard to interpret.

\paragraph{Challenges.} A prime example is given by a set $\overline{\kg}$ with two triples ranked (2,2), MRR $=$ 0.5. After removing their necessary explanations found by three explainability algorithms and retraining the KGE model, we get (2,8), (3,3), (2,4), i.e., ${\text{MRR}}_{1} \simeq$ 0.31, ${\text{MRR}}_{2} \simeq$ 0.33, ${\text{MRR}}_{3} \simeq$ 0.38. The MRR indicates a clear ordering since  ${\text{MRR}}_{1} < {\text{MRR}}_{2}  < {\text{MRR}}_{3}$ (lower MRR is better), but hides details.
The first algorithm changes the rank of one triple by 6, the second changes both ranks by 1 (from 2 to 3), and the third changes the rank of one triple by 2. 
According to the MRR, the second algorithm outperformed the third despite equal cumulative rank changes. This is because the MRR is more sensitive to lower ranks.
Contrarily, the first algorithm performs better due to a single large rank change, outweighing smaller changes in the second. 
Pairing MRR and Hits@2 can clarify such trends because we get Hits@2 $=$ 1, 0, 1 for the three algorithms. Hence, we know that the second algorithm is changing the ranks of both triples, unlike the other two.
Regrettably, coarse resolutions (e.g., Hits@1 or Hits@10) fail to capture these variations as rank changes go undetected. 
This phenomenon occurs in large-scale experiments \cite{zhang2019data,pezeshkpour2019investigating}, where triples with varying ranks are aggregated into $|\overline{\kg}|$.
Here, the Hits@$k$ are ineffective, and the MRR remains constrained by its inherent limitations. 
We summarize our example in Table \ref{tab:example}.

Giving more weight to lower-ranked triples through the MRR is effective for learning KGEs. 
However, we argue that this may not be enough to evaluate explainability algorithms. In fact, a rank change $\Delta$R (the right-hand side of Eq. \eqref{eq: score function nec}) from 1 to 2 could be just as meaningful as one from 2 to 3 since they yield identical shifts in relative performance. Here, the MRRs of 1, 0.5, 0.33 vary non-linearly and reflect the original KGE model rankings rather than the direct impact of the explainability algorithm.


\begin{table}[t]
\centering
\begin{tabular}{ccccc}
\toprule
{Expl. Algo.} & $\rankprime$ & $\Delta$R & {MRR} & {Hits@$\{$1,2,10$\}$} \\
\midrule
1 & (2,\,8) & (0,\,6) & 0.31 & $\{$0, 1, 2$\}$ \\
2 & (3,\,3) & (1,\,1) & 0.33 & $\{$0, 0, 2$\}$ \\
3 & (2,\,4) & (0,\,2) & 0.38 & $\{$0, 1, 2$\}$\\
\bottomrule
\end{tabular}
\caption{Toy example. Two triples change ranks from (2,2) to $\rankprime$. We report rank changes, written as $\Delta$R, MRR, and Hits@$\{$1,2,10$\}$.}
\label{tab:example}
\end{table}

\begin{table}[t]
\centering
\begin{tabular}{lcccc}
\toprule
Algorithm & ML & \text{M}$\Delta$R & MRR & Hits@$1$ [\%]  \\
\midrule
\textbf{{{\kelpie}}} & \textbf{3.92} & \textbf{0.58} & 0.821 & 72 \\
\textbf{{Data Pois.}} & \textbf{1} & \textbf{0.30} & 0.89 & 80 \\
{{\kelpie} (K1)} & 1 & 0.28 & 0.882 & 78 \\
AnyBURL Expl. & 1 & 0.16 & 0.927 & 86 \\
{\criage} & 1 & 0.14 & 0.973 & 96 \\
\bottomrule
\end{tabular}
\caption{Experimental results. ML is the mean explanation length. The rows are sorted by M$\Delta$R. Higher M$\Delta$R and lower MRR and Hits@1 are better. Pareto optimal ML-M$\Delta$R results are bold.}
\label{tab: table results}
\end{table}

\paragraph{How do we improve current evaluations?}

A potential alternative to the MRR is $\text{M}\Delta$R \cite{fuhr2018some}, i.e., the mean of the rank differences $\Delta$R over $\overline{\kg}$. This metric is easily interpretable even when triples in $\overline{\kg}$ have varying ranks, as ranks are not converted to their reciprocals, and measures consistent rank shifts irrespective of triples' starting ranks. On the downside, $\text{M}\Delta$R is sensitive to outliers. For instance, in Table \ref{tab: table results}, \text{M}$\Delta$R is similar for the AnyBURL Explainer and CRIAGE, but we observe this to be due to a single triple in CRIAGE that disproportionately increases its rank with the others mostly unchanged (see Appendix Sec. \ref{apx-sec: experiments details}).
Crucially, M$\Delta$R should be reported together with the explanation length (see Table \ref{tab: table results}). More generally, it is important to integrate both objectives of Prob. \ref{prob: multiobjective}, accounting for effectiveness and conciseness.

A more constraining yet practical solution is to restrict $\overline{\kg}$ to triples with equal ranks $\rank(s,r,o) = k$ and combine MRR and Hits@$k$. 
For this set, the Hits@$k$ count the number of triples shifting their ranks from $k$. 
Also, since all triples start from rank $k$, their MRR can be compared as differences, unlike in the general case where, e.g., a rank change from 1 to 2 and 2 to $+\infty$ yield the same relative rank difference \cite{fuhr2018some}. 
This approach is used for $k=1$ in \cite{rossi2022explaining,bhardwaj2021adversarial,betz2022adversarial,pezeshkpour2019investigating} and in Table \ref{tab: table results}.

To add to these points, we note that these metrics are aggregated and do not indicate \emph{which} triples change ranks. Analyzing this, rather than aggregating triples, may be crucial in applied settings where some target triples are particularly relevant.
Visualizations give this kind of insight (see Appendix Sec. \ref{apx-sec: experiments details}). 
For example, many triples in $\overline{\kg}$ do not change rank across all methods in Table \ref{tab: table results}. Clearly, this approach sacrifices the compactness of tabular data.

While here we focus our discussion on alternative metrics supported by illustrative experiments, we also stress the need to probe algorithms with large-scale experiments in methodological works.

In summary, robust evaluation requires accounting for the nuances of different metrics: relying on a single metric risks drawing misleading conclusions, as no single measure captures all facets of an explanation.

\section{Do Post-hoc Explanations Explain?}
\label{sec: interpretability}

{Are post-hoc explanations of KGC predictions interpretable?} We echo the \enquote{sober look} of \citet{bordt2024statistic}. Namely, we argue that explanations should enable answering interpretative questions, i.e., questions that humans have about machine learning models. Examples in KGC are:
\textbf{Q3.} {Is a chemical reaction $($\text{A}$,$ \text{produces}$,$ \text{B}$)$ ranked worse when $($\text{A}$,$ \text{reacts\_with}$,$ \text{C}$)$ is not observed during training?} 
\textbf{Q4.} {If we train only with a protein-protein interaction $($\text{A}$,$ \text{interacts\_with}$,$ \text{B}$)$, is the enzyme activation $($\text{A}$,$ \text{activates}$,$ \text{C}$)$ still top-ranked?}
\textbf{Q5.} {Does an novel drug-drug interaction $($\text{A}$,$ \text{interacts\_with}$,$ \text{B}$)$ improves the score of $($\text{A}$,$ \text{treats}$,$ \text{C}$)$?}

\textbf{Q3-Q5} differ from \textbf{Q1-Q2} because they situate questions about a statistic, the KGE model score, in user-relevant, real-world applications. It is subtle to understand whether explanations in Sec. \refrange{sec: nec expl}{sec: lat expl} are good answers to \textbf{Q3-Q5}, as by definition, they do not address applied questions.

For necessary and latent explanations $\explstarn$ and $\explstarlplus$, we argue that this may be easier. In fact, $\explstarn$ directly affects a prediction upon its removal from the training set, while $\explstarlplus$ affects it upon its addition. In \textbf{Q3}, a KGE model ranks $($\text{A}$,$ \text{produces}$,$ \text{B}$)$ worse without $\explstarn = ($\text{A}$,$ \text{react\_with}$,$ \text{C}$)$. In \textbf{Q5}, training upon adding $\explstarlplus = ($\text{A}$,$ \text{interacts\_with}$,$ \text{B}$)$ to the training set improves the score of $($\text{A}$,$ \text{treats}$,$ \text{C}$)$.

Sufficient explanations become challenging if extracted by proxy as per Def. \ref{def: suf expl}, i.e., by \enquote{transferring} them to $\mathcal{C}$. This approach makes experiments with KGE models feasible, but it introduces uncertainty in queries such as \textbf{Q4}.
This phenomenon is evident in the example of Sec. \ref{sec: suff expl}, where a path between the subject and object of a prediction serves as a better sufficient explanation than a single triple extracted with Def. \ref{def: suf expl}.
GNN-based explainability algorithms \cite{ma2024kgexplainer} can help overcome these limitations. However, as seen in Sec. \ref{sec: suff expl}, it is important to consider data contamination.

Additionally, it is important to highlight that while explanations may inform us about a KGE model's prediction, they do not reveal real-world causal relationships \cite{molnar2022general}. E.g., in \textbf{Q5}, a drug-drug interaction does not imply a better treatment outcome; this must be tested in laboratory experiments.

Our discussion underscores the need to parse explanations through human evaluation, as omissions may lead to explainability claims without interpretation. Fortunately, this is an emerging trend \cite{sudhahar2024experimentally,huang2024foundation,ma2024kgexplainer,chang2024path}.  We also discuss alternative views in Appendix Sec. \ref{apx-sec: alternative views}.

\section{Conclusion}

We proposed a taxonomy for post-hoc explanation in KGC, tackling literature gaps.
We characterized post-hoc explanations with multi-objective optimization, unifying explanation types, 
and discussed evaluation protocols for explainability algorithms and the interpretability of explanations.

Future methodological research directions, in alignment with our work, include studying latent explanations, e.g., by advancing probabilistic generative models for KGs, refining proxies for sufficient explanations, and developing explainability algorithms that use fewer resources while enabling the exploration of larger search spaces.

\bibliographystyle{unsrtnat}
\bibliography{main}

\clearpage
\newpage
\appendix

\section{Details about explainability algorithms}
\label{apx-sec: explainability algorithms}

\subsection{CRIAGE \cite{pezeshkpour2019investigating}}
\label{apx-ssec: criage}

CRIAGE (Completion Robustness and Interpretability via Adversarial Graph Edits) estimates the effect of post-hoc explanations on the predictions of a KGE model as follows.

For a necessary explanation, the difference between the KGE model loss before and after explanation removal is approximated by truncating its Taylor expansion at the first order with respect to the embeddings. This approach allows the authors to derive a closed-form expression for the difference $\scorefnprime(s_x,r_x,\hat{o}_x) - \scorefn(s_x,r_x,\hat{o}_x)$ (an effectiveness function akin to Eq. \eqref{eq: score function nec}). Then, the most effective neighboring triple to $(s_x,r_x,\hat{o}_x)$ is found through exhaustive search.

For adversarial attacks (aimed at both increasing or decreasing the prediction score), using brute force to find the most effective triple to add would not work. In fact, constraining the explanation objective to $o = \hat{o}_x$ leaves room for its subject-relation tuple to vary across $|\mathcal{R}| |\entities|$ values. To alleviate the complexity of this search, the authors translate the search problem into a continuous optimization problem. Specifically, they use gradient descent to find the most effective unobserved triple, optimizing its vector representation in the embedding space. Once the optimal vector is found, its corresponding $(s,r)$ pair is retrieved using a decoder (an \enquote{Inverter Network}) specifically trained for such a purpose.

CRIAGE is reprogrammed to find $\mathcal{C}$-sufficient explanations in \citet{rossi2022explaining}. The method used is the same as that for necessary explanations. However, the score difference is now calculated for target triples $(c,r_x,\hat{o}_x) \in \kg_{\mathcal{C}}$ upon the addition of a triple among those connected to $s_x$. The best triple is chosen through exhaustive search.

\subsection{AnyBURL Explainer \cite{betz2022adversarial}}
\label{apx-ssec: anyburl explainer}

The AnyBURL Explainer is a black-box method to find explanations from predictions of AnyBURL  \cite{meilicke2019anytime}, a symbolic algorithm for learning logical rules for KGC.

AnyBURL samples paths from a KG and transforms them into logical rules. In particular, triples in the KG are abstracted into facts, i.e., statements symbolically denoted as $\mathsf{r}(s,o)$. Here, the subject $s$ is connected to the object $o$ through the predicate $\mathsf{r}$. Facts whose corresponding triples are connected in KG paths, e.g., $($\text{Alice}, \text{lives\_in}, \text{Netherlands}$)$ and $($\text{Netherlands}, \text{language\_is}, \text{Dutch}$)$, are used to learn rules such as \text{lives\_in}(X, A) and \text{language\_is}(A, Y), relating the variables $X$, $A$, and $Y$ with the predicates \text{lives\_in} and \text{language\_is}. This enables prediction. In this example, it would consist of inferring that Alice, a particular instance of $X$, speaks Dutch, an instance of $Y$, due to the conjunction of the two rules.

A similar mechanism is used to find explanations. Necessary candidate explanations are represented by rules of triples connected to the predicted triple that, when instantiated, enable one to make such a prediction. The best necessary explanation is selected based on a confidence score. Intuitively, this is the ratio between the number of instantiations for which a prediction is true, divided by the total number of times the candidate rule can be instantiated.

Adversarial attacks (aimed at worsening predictions) are carried out by modifying the extracted explanations. In particular, for a necessary explanation $(s, r, o)$, if $s = s_x$ (equivalently, $o = \hat{o}_x$), the object $o$ (or equivalently, $s$) is swapped with an entity to produce an unobserved triple in the KG.

\subsection{Data Poisoning \cite{zhang2019data}}
\label{apx-ssec: data poisonining}

Data Poisoning finds explanations belonging to two categories. The first explanations are deemed by the authors \enquote{direct attacks} and consist of triples $(s,r,o)$ whose subject is $s=s_x$. The second explanations are called \enquote{indirect attacks}. They are triples with entities within 1 hop from $s_x$ but not necessarily connected to it. These two search spaces correspond to those represented in Fig. \ref{fig: nec-expl-fig}(b).

For direct attacks, necessary explanations are extracted as follows. First, the partial derivative of $\scorefn(s_x, r_x, \hat{o}_x)$ with respect to the embedding of $s_x$ is computed and multiplied by a small scalar, the \enquote{perturbation step}. This product, $\Tilde{\boldsymbol{\epsilon}}$,  represents the \enquote{fastest direction [to decrease the prediction score]}. Then, the difference $\scorefn(s = s_x, r, o) - \lambda \Tilde{\Phi}_{\boldsymbol{\theta}}(s = s_x, r, o)$ is calculated for all triples connected to the prediction with $s=s_x$. Here, $\lambda$ is a scalar parameter and $\Tilde{\Phi}_{\boldsymbol{\theta}}(s = s_x, r, o)$ is the score of $(s = s_x, r, o)$ obtained when shifting the embedding of $s_x$ by $\Tilde{\boldsymbol{\epsilon}}$. The $M$ triples maximizing this difference are taken as necessary explanations. The intuition behind this construction is that such triples maximize $\scorefn(s = s_x, r, o)$ and therefore, they are relevant for the KGE model to infer $(s_x, r_x, \hat{o}_x)$. Additionally, they yield small values of $\Tilde{\Phi}_{\boldsymbol{\theta}}$, i.e., they are not relevant for the KGE model if it were to make a wrong prediction.

An analogous strategy is used for adversarial attacks. Here, the goal is to worsen a prediction $(s_x, r_x, \hat{o}_x)$. Therefore $\lambda \Tilde{\Phi}_{\boldsymbol{\theta}}(s = s_x, r, o) - \scorefn(s = s_x, r, o)$ is to evaluate the effectiveness of explanations.

Indirect attacks are more complex. Without going into extensive details, the authors develop a heuristic to estimate how the perturbation of a removed triple would propagate to $(s_x,r_x,\hat{o}_x)$ passing through other triples. Such a heuristic is detailed in Algorithm 1 of \citet{zhang2019data}. 

Data Poisoning is reimplemented in \citet{rossi2022explaining} to find sufficient explanations. Here, instead of computing $\Tilde{\boldsymbol{\epsilon}}$ at $s_x$, the partial derivative of the KGE score is computed at $c$ for each $(c,r_x,\hat{o}_x)$. Triples in the neighborhood of $(s_x,r_x,\hat{o}_x)$ are linked to every $(c,r_x,\hat{o}_x)$ with a $s_x$-to-$c$ swap. Then, the triple leading to the greatest score improvement of $(c,r_x,\hat{o}_x)$ is taken as a $\mathcal{C}$-sufficient explanation.

\subsection{Kelpie \cite{rossi2022explaining}}
\label{apx-ssec: kelpie}

Kelpie extracts necessary and sufficient explanations of variable length using three main modules: a \enquote{Pre-Filter}, a \enquote{Relevance Engine}, and an \enquote{Explanation Builder}.

The \enquote{Pre-Filter} reduces the search space $\mathcal{X}$, making computations with Kelpie more efficient. For a prediction $(s_x, r_x, \hat{o}_x)$, it restricts the search space using a top-$K$ policy ($K =$ 10, 20, 30). This policy considers triples $(s = s_x, r, o)$ or $(s, r, o = s_x)$ and selects the ones with the smallest shortest non-oriented shortest path between $s$ and $\hat{o}_x$, or between $o$ and $\hat{o}_x$, respectively.

The \enquote{Relevance Engine} estimates the effect of removing a triple (a candidate necessary explanation) or adding it to a target triple (a candidate sufficient explanation). It does so by retraining the KGE model. However, since retraining from scratch would be computationally prohibitive, it adopts a \enquote{post-training} method. For necessary explanations, this consists of retraining the KGE model after removing a candidate explanation (a set of triples) while keeping all embeddings frozen as in the original training, except for those of the entities and relations in triples connected to $s_x$. For sufficient explanations, the approach is analogous, but the retrained embeddings are only those of triples connected to target entities $c$. The difference in rank of $(s_x, r_x, \hat{o}_x)$ (respectively, $(c, r_x, \hat{o}_x)$ for sufficient explanations) before and after the \enquote{post-training} is calculated as in Eq. \eqref{eq: score function nec} (respectively, Eq. \eqref{eq: score function suf} with an extra normalization factor $\rankprime(c,r_x,\hat{o}_x) - 1$). Such a difference is used to measure the effectiveness of explanations.

The \enquote{Explanation Builder} constructs explanations for evaluation by the \enquote{Relevance Engine} using Adaptive Simulated Annealing \cite{ingber1989very}, enabling the creation of explanations with multiple triples without extensive search. The routine first explores single-triple explanations and computes their effectiveness with the \enquote{Relevance Engine}. If an optimal explanation is found (where optimality is measured by the effectiveness surpassing a predefined threshold), the search stops. Otherwise, if no single-triple explanation is optimal, explanations of length two are evaluated by the \enquote{Relevance Engine} in order of preliminary relevance, computed as the sum of the effectiveness of their individual triples. This process continues up to explanations made of four triples.

\section{An example for $\mathcal{C}$-sufficient explanations}
\label{apx-sec: more suff expl}

\begin{figure*}[htpb]
    \centering
    \includegraphics[width=0.80\textwidth]{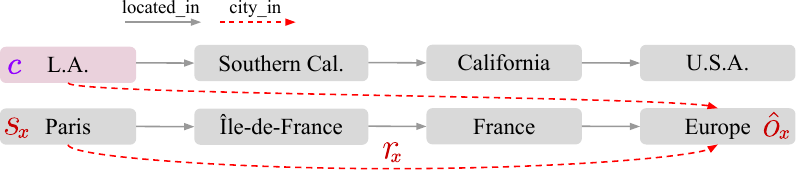}
    \caption{Example KG. The target set is made up of a single triple, $\mathcal{C} = \{ c \} = \{ \text{L.A.} \} $, where the $s_x$-to-$c$ swap happens changing $(s_x,r_x,\hat{o}_x) = (\text{Paris}, \text{city\_in}, \text{Europe})$ to $(c,r_x,\hat{o}_x) = (\text{L.A.}, \text{city\_in}, \text{Europe})$.}
    \label{fig: suppl-fig-expl}
\end{figure*}

In Fig. \ref{fig: suppl-fig-expl}, we draw the KG discussed in the example of Sec. \ref{sec: suff expl}. Here, the prediction is $(s_x, r_x, \hat{o}_x)$  $=$ $($\text{Paris}$,$ \text{city\_in}$,$ \text{Europe}$)$. A sufficient explanation is $($\text{Paris}$,$ \text{located\_in}$,$ \text{Île-de-France}$)$, $($\text{Île-de-France}$,$ \text{located\_in}$,$ \text{France}$)$, $($\text{France}$,$ \text{located\_in}$,$ \text{Europe}$)$. However, its $\mathcal{C}$-sufficient explanation would be $($\text{Paris}$,$ \text{located\_in}$,$ \text{Île-de-France}$)$, which gets connected to $($\text{L.A.}$,$ \text{located\_in}$,$ \text{Southern California}$)$.

\section{Negative latent explanations}
\label{apx-sec: latent explanations}

Analogously to positive sufficient explanations in Sec. \ref{sec: suff expl}, negative sufficient explanations are unobserved triples \enquote{likely with respect to $q$}. However, they are opposite in their scope, as their effectiveness is measured by computing how much the rank of a prediction increases, i.e., how much its score worsens.

We define them as follows.

\begin{definition}[\emph{Negative Latent Explanation}]
\label{def: neg lat expl}
A negative latent post-hoc explanation $\explstarlminus$ for a prediction $(\subject_x, \relation_x, \hat{\object}_x)$ with rank $\rank(\subject_x, \relation_x, \hat{\object}_x) < |\entities|$ is the smallest non-empty set of unobserved triples in $\Omega \setminus \kg_\train$ \enquote{likely with respect to $q$} that, if added to $\kg_\train$ and upon retraining the $\kge$ model on $\kg_\train \cup \explstarlplus$, maximally increases $\rank(\subject_x, \relation_x, \hat{\object}_x)$.
\end{definition}

Negative latent explanations are solutions of Problem 1 with steps \textbf{(i)-(iii)} as for positive ones. Here, however, the goal is to calculate the increase in the rank of a prediction. Therefore, to measure explanation in \textbf{(iii)}, we can take, e.g., $-\Psi$, with $\Psi$ as in Eq. \eqref{eq: score function nec}.

Their interpretation becomes clearer with an example. We can understand them as the smallest set of triples that, if observed, would \enquote{maximally invalidate} a prediction made by the KGE model. Take a prediction such as $(\text{France}, \text{located\_in}, \text{Asia})$. In this case, triples like $(\text{France}, \text{neighbor\_of}, \text{Italy})$ or $(\text{France}, \text{neighbor\_of}, \text{Germany})$ have the opposite effect than a positive explanation in Def. \ref{def: pos lat expl}: they worsen the prediction by decreasing its score and, thus, increasing its rank.

\section{Details about the experiments}
\label{apx-sec: experiments details}

\subsection{Training ComplEx}

The numerical results in Sec. \ref{sec: experiments} are obtained by training ComplEx \cite{trouillon2016complex} using the code from \citet{rossi2022explaining}, which is primarily based on the implementation of \citet{lacroix2018canonical}. ComplEx is a KGE model that solves KGC through low-rank matrix factorization with complex embeddings. We train it on FB15k-237 \cite{toutanova2015observed}, a popular KGC dataset derived from FB15k \cite{bordes2013translating} by filtering out data leaks between the evaluation and test sets. FB15k-237 contains 310,116 triples, 14,541 entities, and 237 relations.

\begin{figure*}[htpb]
    \centering
    \includegraphics[width=0.65\textwidth]{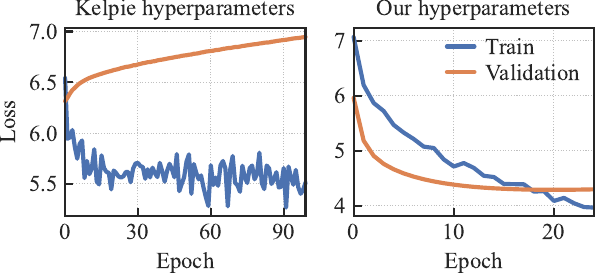}
    \caption{Left panel: Training and Validation loss of ComplEx with the hyperparameters suggested by \citet{rossi2022explaining}. Right panel: Losses of ComplEx with our hyperparameters.}
    \label{fig: complex-training-fig}
\end{figure*}

Initially, we trained ComplEx using the hyperparameters suggested in the Kelpie repository\footnote{\texttt{https://github.com/AndRossi/Kelpie}} \cite{rossi2022explaining} by its authors. This setup produced good performance in terms of Test and Validation Hits@$k$ ($k=$1, 10) and MRR, as shown in Tab. \ref{tab:complex_performance} (\enquote{Kelpie} column). However, upon closer inspection, we observed that the model validation loss increased during training (see Fig. \ref{fig: complex-training-fig}). This behavior was not reflected in the Hits@$k$ or MRR metrics but was evident in the Negative Log-Likelihood loss, which ComplEx is trained on.

Intuitively, this discrepancy arises because metrics like MRR and Hits@$k$ focus on ranks, which discretize differences in the KGE score into unit-sized intervals. Differences in the score are instead captured by the loss. This metric, being sensitive to score differences, may explode if large score differences are observed.

This observation does not affect the experimental results in \citet{rossi2022explaining}, as the KGE model's performance is assessed using MRR and Hits@$k$. However, we decided to retrain ComplEx with a new set of hyperparameters, as reported in our code repository, to make the model viable for scenarios where the KGE score matters, e.g., confidence thresholds or score-based filtering. With this new set of hyperparameters, ComplEx still performs well in terms of MRR and Hits@$k$ as shown in Tab. \ref{tab:complex_performance} (\enquote{Ours} column). Additionally, the validation loss decreases as shown in Fig. \ref{fig: complex-training-fig}. We use the new hyperparameters for our explainability experiments.

\subsection{Running the explainability algorithms}

We randomly select 50 triples among those top-ranked by ComplEx. To ensure reproducibility, we provide these triples in our code repository. Along with the top-ranked triples, we include detailed commands and instructions to replicate the results in Tab. \ref{tab: table results}. CRIAGE, Data Poisoning, and Kelpie were executed using their default configurations as implemented in \citet{rossi2022explaining}. Similarly for the AnyBURL Explainer \cite{betz2022adversarial}, which is bundled with AnyBURL v22, available through the AnyBURL repository\footnote{\texttt{https://web.informatik.uni-mannheim.de/AnyBURL/}}.

Following the approach used in Kelpie, we extract necessary explanations for the 50 top-ranked sampled triples. Then, we remove them simultaneously from the training set and retrain ComplEx on this modified set.

\begin{figure*}[htpb]
    \centering
    \includegraphics[width=0.8\textwidth]{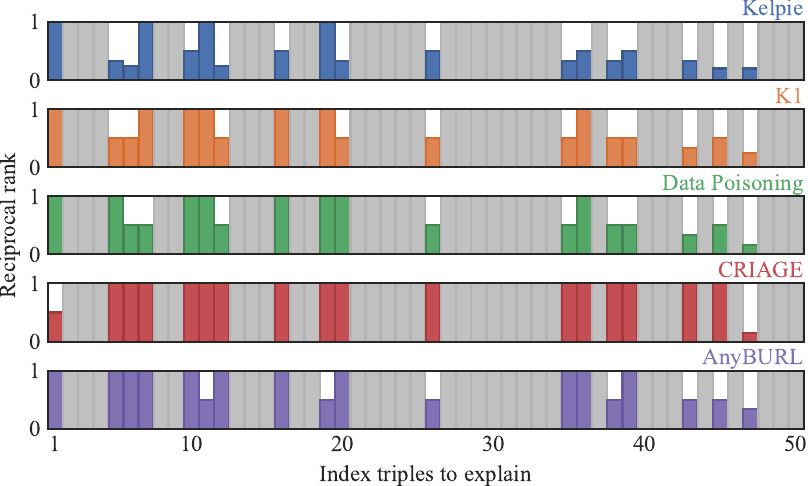}
    \caption{Reciprocal ranks of all 50 random test triples. Triples that do not change ranks after the explanation's removal and KGE model's retraining are colored in gray.}
    \label{fig: suppl-fig-MRR}
\end{figure*}

\begin{figure*}[htpb]
    \centering
    \includegraphics[width=0.8\textwidth]{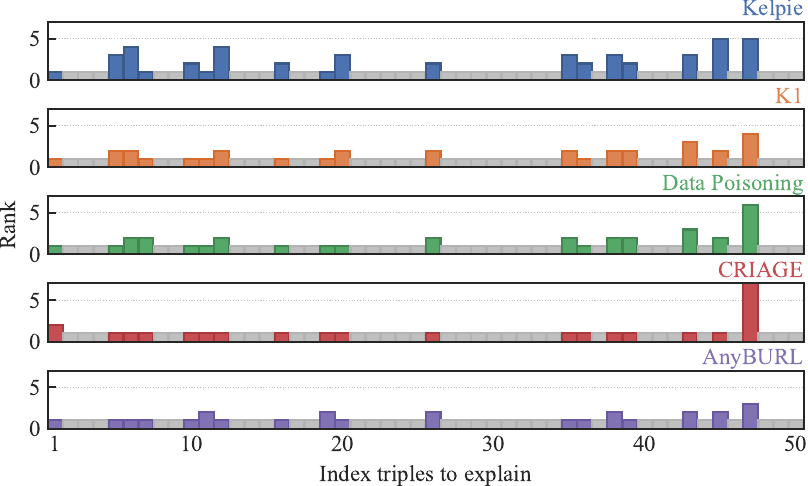}
    \caption{Ranks of all 50 random test triples. Triples that do not change ranks after the explanation's removal and KGE model's retraining are colored in gray.}
    \label{fig: suppl-fig-MR}
\end{figure*}

As discussed in Sec. \ref{sec: experiments}, in Fig. \refrange{fig: suppl-fig-MRR}{fig: suppl-fig-MR}, we display bar plots showing the reciprocal ranks and ranks for each explainability method across all 50 drawn triples.

\begin{table*}[htpb]
\begin{center}
\begin{tabular}{lcc}
\toprule
Metric & Ours & Kelpie \\
\midrule
Validation Hits@1 & 0.233 & 0.275 \\
Validation Hits@10 & 0.484 & 0.561 \\
Validation MRR & 0.316 & 0.37 \\
Test Hits@1 & 0.227 & 0.271 \\
Test Hits@10 & 0.478 & 0.558 \\
Test MRR & 0.311 & 0.365 \\
\bottomrule
\end{tabular}
\end{center}
\caption{Validation and Test performance metrics for our ComplEx hyperparameters are under \enquote{Ours}. Those of \citet{rossi2022explaining} are under \enquote{Kelpie}. Numbers are rounded to the third decimal digit.}
\label{tab:complex_performance}
\end{table*}

\subsection{Comments on the interpretability of M$\Delta$R}

Note that M$\Delta$R, the mean of the rank differences $\Delta$R over a test set $\overline{\kg}$, is the same as $\Delta$MR, the difference in the arithmetic mean over all individual ranks.
Indeed,
\begin{align}
    \text{M}\Delta\text{R} & = 
    \frac{1}{|\overline{\kg}|} \left( \sum_{(s,r,o) \in \overline{\kg}} \rankprime(s,r,o) - \rank(s,r,o) \right) \\
    &= \frac{1}{|\overline{\kg}|} \left( \sum_{(s,r,o) \in \overline{\kg}}  \rankprime(s,r,o) \right) -\frac{1}{|\overline{\kg}|} \left(  \sum_{(s,r,o) \in \overline{\kg}}  \rank(s,r,o) \right) \\
    &= \text{MR}_{\btheta\boldsymbol{'}} - \text{MR}_{\btheta} = \Delta\text{MR} \,.
\end{align}

Although MR has the unusual property for an evaluation metric of indicating better performance for lower values, it is easily interpretable and weighs rank changes consistently on the whole scale \cite{fuhr2018some}.
When evaluating rank changes in M$\Delta$R, we use the new ranks as the subtrahend and the original ranks as the minuend. Specifically, we compute $\text{MR}_{\btheta\boldsymbol{'}} - \text{MR}_{\btheta}$ instead of $\text{MR}_{\btheta} - \text{MR}_{\btheta\boldsymbol{'}}$. As a result, we recover a metric where higher values indicate better performance.

\section{Alternative Views}
\label{apx-sec: alternative views}

Several viable perspectives alternative to ours warrant consideration.

First, \emph{explanations extend beyond triple-based representation, they are multifaceted}. 
Textual explanations, for instance, give more intuitive and accessible insights for specific audiences. Large Language Models could generate potential explanations to complement or replace triples \cite{rasheed2024knowledge}. 
For example, if we take a prediction $($\text{A}$,$ \text{produces}$,$ \text{B}$)$ in a medical context, a textual explanation like \enquote{Compound A reacts with Enzyme X to produce Compound B, a key process in metabolic pathway Y} may better resonate with medical professionals.

Second, \emph{interpretability is user-dependent}. 
The utility of an explanation depends on user needs, domain-specific constraints, and task-specific requirements. 
For example, a data scientist might prefer graph-based explanations, whereas a medical professional may value concise, domain-specific textual ones.
Addressing these needs may require evaluations tailored to the unique demands of diverse use cases, beyond those discussed. For example, it could be relevant to embed desirable properties of explanations, like their alignment with users' prior knowledge \cite{nauta2023from}.

\end{document}